%% file: paper.tex
\RequirePackage{fix-cm}
\documentclass[twocolumn]{svjour3}          %

\smartqed  %
\usepackage{graphicx}

\usepackage{times}
\usepackage{epsfig}
\usepackage{graphicx}
\usepackage{amsmath}
\usepackage{amssymb}
\usepackage{xspace}
\usepackage{rotating}
\usepackage{units}
\usepackage{multirow}
\usepackage{algorithm}
\usepackage{lettrine}

\usepackage{color}

\usepackage[pagebackref=false,breaklinks=true,letterpaper=true,colorlinks,bookmarks=false]{hyperref}

\renewcommand{\paragraph}[1]{{\medskip \noindent \bf #1}}

\def\qu{\boldsymbol{q}}
\def\b{\boldsymbol{b}}
\def\ftqu{\mathcal{Q}}
\def\ftb{\mathcal{B}}
\def\f{\boldsymbol{f}}

\def\s{\boldsymbol{s}}

\def\eone{\boldsymbol{e}_1}

\def\O{{\mathcal O}}

\def\eg{\emph{e.g.}\xspace}
\def\ie{\emph{i.e.}\xspace}
\def\etc{\emph{etc}\xspace}
\def\wrt{\emph{w.r.t.}\xspace} 
\def\etal{\emph{et al.}\xspace}

\def\ourmethod{Circulant Temporal Encoding\xspace}
\def\ourbaseline{Mean-MultiVLAD\xspace}
\def\BSL{MMV\xspace}  %
\def\OUR{CTE\xspace}  %

\def\trecvid{\textsc{Trecvid}\xspace}
\def\ccweb{\textsc{CCweb}\xspace}
\def\ccwebdist{\textsc{CCweb+100k}\xspace}

\def\aligngui{\textsc{AlignGUI}\xspace}

\def\sect#1{Section~\ref{sec:#1}}

\begin{document}

\title{Circulant temporal encoding for video retrieval and temporal alignment}

\author{
Matthijs Douze 
\and J\'er\^ome Revaud 
\and Jakob Verbeek
\and Herv\'e J\'egou 
\and Cordelia Schmid 
}
\institute{\email{fistname.lastname@inria.fr}           %
}
\date{Received: date / Accepted: date}

\maketitle
\thispagestyle{empty}

\begin{abstract}
We address the problem of specific video event retrieval.
Given a query video of a specific event, \eg, a concert of Madonna, the goal is to retrieve other videos of the same event that temporally overlap with the query. 
Our approach encodes the frame descriptors of a video to jointly represent their appearance and temporal order. It exploits the properties of circulant matrices to efficiently compare the videos in the frequency domain. This offers a significant gain in complexity and accurately localizes the matching parts of videos. The descriptors can be compressed in the frequency domain with a product quantizer adapted to complex numbers. In this case, video retrieval is performed without decompressing the descriptors.

We also consider the temporal alignment of a set of videos. 
We exploit the matching confidence and an estimate of the temporal offset computed for all pairs of videos by our retrieval approach. 
Our robust algorithm aligns the videos on a global timeline by maximizing the set of temporally consistent matches.  
The global temporal alignment enables synchronous playback of the videos of a given scene.
\end{abstract}

\input{intro.tex}

\input{method.tex}
\input{indexing.tex}

\input{video_align.tex}

\input{dataset.tex}

\input{experiments.tex}

\input{conclusion.tex}

\bibliographystyle{ieee}
\bibliography{egbib}

\end{document}

%% file: intro.tex
\section{Introduction}
\label{sec:intro}

\lettrine{T}{his} paper presents our circulant temporal encoding (CTE) approach, to  determine whether two videos represent the same event and share a temporal and visual overlap. 
It encodes all the frame descriptors of a video into a temporal
representation and exploits the properties of circulant matrices to
compare videos in the frequency domain. The result of the comparison
is a similarity score and a time shift that temporally aligns the two
videos.

We first use CTE to address the problem of video copy detection: given a query video, the goal is to 
retrieve exact or near duplicate clips or sub-clips from a large video collection. 
Examples of applications are automatic copyright enforcement, 
identification of advertisement videos 
or automatic organization of video collections. %
We show that our approach can retrieve different videos of the 
same event filmed from different viewpoints. For instance, CTE can recognize that
several amateur videos shot by different persons attending the same concert do correspond.
Indexing this type of video material on-line and in archives  enables to mine video data in large
archives that are often indexed with sparse and noisy keywords.   

CTE can also be applied to automatic temporal alignment of a set of videos of the same  event, 
in order to play back the videos synchronously. 
For example, home users who have shot tens of videos of an event such as a
wedding or a sports event would like to combine them into a
single movie of the event. 
Professional users want to organize a collection of videos from
different viewpoints of the same scene, shot under conditions that
prevented them to gather appropriate metadata. Video collections of an
event such as a theatrical performance retrieved from external sources
such as YouTube or other video sharing sites, can be edited into a single
movie.

A preliminary version of this work, introducing CTE for the purpose of large-scale video retrieval, appeared in~\cite{CTEPaper}.
The core contribution of CTE is to  
jointly encode in a single vector the appearance information
at the image level and the temporal sequence of
frames, such that it enables an efficient search scheme
that avoids the exhaustive comparison of frames. 
Extensive experiments on two video copy detection benchmarks in~\cite{CTEPaper}
have shown that CTE improves over the state of the art with respect
to accuracy, search time and memory usage, and allows as well to retrieve
video of specific instances of events despite important viewpoint changes
in a large collection of videos.

Although we briefly mentioned in~\cite{CTEPaper} the possibility of
aligning multiple videos with CTE, we did not describe the 
approach in detail and there was no experimental evaluation.
In this paper, we introduce a new  dataset for global video alignment. 
It consists of temporally aligned videos of a rock climbing session, captured with different cameras including mobile phones, camcorders and head-mounted cameras. 
We also annotate the ground-truth for the temporal alignment of 163 clips of the \emph{Madonna in Rome} event from 
the publicly available EVVE dataset \cite{CTEPaper}.
Furthermore, we present a robust temporal alignment algorithm that
takes as input pairwise video matches and their estimated temporal
shift, and produces a global temporal alignment of the videos. This
alignment process is evaluated using the ground-truth alignment of
the \emph{Madonna} and \emph{Climbing} datasets.

The rest of this paper is organized as follows. 
Section~\ref{sec:related} reviews related work on video retrieval and alignment.
Section~\ref{sec:method} describes our  temporal
circulant encoding technique,  Section~\ref{sec:indexing} presents
our indexing strategy and  Section~\ref{sec:videoalign} introduces our robust 
video alignment method. 
Section~\ref{sec:dataset} introduces the datasets used in the experiments and the related evaluation measures. 
Experimental results in Section~\ref{sec:experiments} demonstrate the
excellent performance of our
approach for video retrieval and alignment.
Finally, we present our conclusions in Section~\ref{sec:conclusion}.

\section{Related work}
\label{sec:related}

We review related work on video retrieval and alignment in Sections 
\ref{sec:related_retrieval} and \ref{sec:related_alignment} respectively.

\subsection{Video retrieval}
\label{sec:related_retrieval}

Searching for specific events is related to video copy
detection~\cite{LTJ07} and event category
recognition~\cite{OVS13}, but there are substantial   
differences with both tasks. The goal of video copy detection   
is to find videos that are distorted versions of the query video, \eg, by compression, cam-cording or  picture-in-picture transformations.
Detecting event \emph{categories} requires a classification approach
that captures the large intra-class variability in semantically
defined events like \emph{brush hair} in HMDB51~\cite{kuehne2011hmdb},
\emph{birthday party} in Trecvid MED~\cite{2014trecvidover}, or \emph{high jump} in 
UCF101~\cite{soomro2012ucf101}. Such datasets provide 
a large pool of positive and negative examples to train the classifier. 
In contrast, the method introduced in this paper is tailored to specific event
retrieval, that is, finding videos of a particular event instance. It,
thus, needs to be flexible enough to handle significant viewpoint
changes, but should nevertheless produce a precise alignment in time.  
Retrieval is performed given one video query,  \ie without training data.

This is substantially different from approaches aiming at discovering high-level correspondences between semantic actions in different videos \cite{chu12eccv,hoai12aistats,xiong12mdm}. 
These methods discover common temporal patterns between videos in an unsupervised
fashion, and allow the matched subsequences to have varying length. Both \cite{hoai12aistats,xiong12mdm} rely on high-level descriptors and flexible alignment methods based on discriminative clustering in order to match different instances of the same action classes. The discriminative clustering approach characterizes common subsequences of videos (clusters) using separating hyperplanes, which can only be reliably estimated from lager collections of videos, and cannot be readily used to measure the similarity between two videos.
The branch-and-bound technique of \cite{chu12eccv} can in principle be adapted to force matching subsequences to be of the same length. The bounds used in their work, however, rely on representing a sequence by accumulating individual frame feature histograms. This temporal aggregation makes precise temporal alignment particularly problematic with their approach.
Our goal is to retrieve videos of the same specific event instance as the query video. 
In this context, a precise temporal alignment is well-defined and feasible. 
We exploit this property to tackle difficult viewpoint changes with little discriminant motion.

Many techniques for video retrieval represent a video as a set
of descriptors extracted from frames or keyframes~\cite{DJSP10,Karpenko2011,Song2011}.
Searching in a collection is performed by comparing the query descriptors
with those of the dataset. Then, temporal constraints are 
enforced on the matching descriptors, by \eg, partial alignment~\cite{YC09}
or classic voting techniques, such as temporal Hough transform~\cite{DJSP10},
which was popular in the \trecvid video copy detection task~\cite{SOK06}. 
Such approaches are costly, since 
all frame descriptors of the query must be compared to those
of the database before performing the temporal verification. 

In contrast, the technique proposed in this paper measures the
similarity between two sequences for all possible
alignments at once. 
Frame descriptors are jointly encoded in the 
frequency domain, where convolutions cast into efficient element-wise multiplications. 
This encoding is combined with frequency pruning to
avoid the full computation of all cross-similarities between the frame
descriptors. 
The comparison of sequences is improved by a regularization in the frequency domain. 
Computing a matching score between videos  requires only component-wise operations
and a single one-dimensional inverse Fourier
transform, avoiding the reconstruction of the descriptor in the
temporal domain. 

Similar techniques have been used for 
registration or watermark detection. However, they are usually
applied to the raw signal such as image
pixels~\cite{B92,HCMB12} or audio waveforms~\cite{KDHM99}. 
Transforming a multi-dimensional
signal to the Fourier domain was shown useful to speed up object detection~\cite{Dubout2012,henriques13iccv}.

We optimize the tradeoff between search quality, speed and memory consumption using  product quantization~\cite{JDS11},
which we extend to complex vectors in order to compare our descriptors in the compressed Fourier domain.

\subsection{Precise video alignment}
\label{sec:related_alignment}

Existing temporal video alignment techniques for videos of the same specific event are limited to constrained environments, or impose strong requirements on the video capturing process. Below, we review the most important existing alignment techniques.

Genlocking cameras consists in triggering the shutter of the cameras 
  by a common clock signal~\cite{grimagesetup}. This ensures that the
  frames are perfectly synchronized, which is useful for the 3D~reconstruction of fast-moving scenes~\cite{franco:inria-00349103}.
A movie clapperboard or a light flash~\cite{jiang3d} can be used to
recognize a common time instant in the different videos. This approach
also imposes constraints on the capturing process, even though the problem is simplified. 

When the cameras have approximately the same viewpoint, the optimization of an error criterion between the image pixels can be used to compute a spatio-temporal model mapping one video to another~\cite{CaspiPAMI02}. 
In another work~\cite{TinneCVPR04}, the alignment is obtained by explicitly matching scene points and using geometrical constraints on the matches to spot consistent frames. This is extended in~\cite{EvangelidisPAMI13} to a spatio-temporal model with a non-constant time shift, which can be used for example to match videos of different passages through the same street. 
Another work along the same lines is VideoSnapping~\cite{wang2014videosnapping}, %
which focuses on adjusting the speed of one video to match that of a
reference one. Similar to our method, VideoSnapping also addresses the
alignment of multiple videos using a minimum spanning tree. In both \cite{EvangelidisPAMI13} and~\cite{wang2014videosnapping}, the temporal model is more general than ours: instead of a constant time shift, they estimate a dynamic time warping (DTW). This is much harder because the number of parameters is proportional to the length of the movies, and restricts these methods to relatively high-quality footage with clean keypoints (\eg strong corners on buildings, square objects). In contrast, our method can deal with less restrictive settings such as low-quality videos from YouTube or videos that are static for long periods, as shown on Figure~\ref{fig:pairwisematches}. 

The audio produced along  with the different cameras can be aligned more easily, as it is a one dimensional signal~\cite{Hasler2009,rockalign}. 
However, this can be exploited only if the scene is sufficiently small to avoid a drift due to the
relatively slow propagation of sound in the air. Moreover, audio may not be distinctive enough (\eg, if the audio track is dominated by motor sounds), or its quality may be too poor.  

In contrast to these approaches, our approach does not impose constraints on the capturing process. Therefore, unlike genlocking or clapperboard-like techniques, our approach is applicable to videos of an event acquired in an uncontrolled and uncoordinated  manner.
Compared to audio-based approaches, our approach is more robust since we rely on the richer visual signal.

\newcommand{\igcross}[1]{\framebox{\includegraphics[width=0.3\linewidth]{figs/match_matrix/#1.png}}}
\begin{figure}
\begin{center}
\igcross{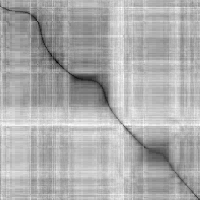}%
\igcross{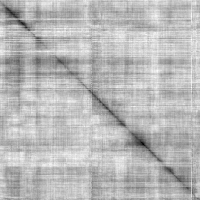}%
\igcross{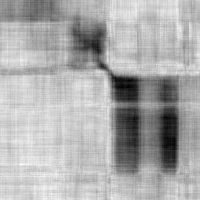}
\end{center}
\caption{\label{fig:pairwisematches}
        Matrices of frame matching scores for pairs of videos to align. Each matrix element is the matching score between a query frame (line) and a database frame (column). Darker entries correspond to higher scores. Left: sequences with non-constant time shift (we do not consider this case); middle: constant time shift, strong signal; right: constant time shift, weak signal (the focus of this paper). 
}
\end{figure}

%% file: method.tex
\section{Video matching  using circulant temporal encoding}
\label{sec:method}

\def\q{}

This section describes the algorithm for video matching. 
We first present the video descriptors in Section \ref{sec:descriptors}, and the general matching formulation in Section \ref{sec:matching}. We then present the efficient circulant temporal encoding technique in Section \ref{sec:circ}, and a robust regularized version of it in Section \ref{sec:regmetric}
Finally, in Section \ref{sec:boundetect} we discuss how the matching subsequences are extracted.

\subsection{Video frame descriptors}

\label{sec:descriptors}

We represent a video by a sequence of high-dimensional frame
descriptors. 
The representation is obtained in the following three steps.

\paragraph{Pre-processing.}
All videos are mapped to a common format, by sampling them at a fixed
rate of 15~fps and resizing them to a maximum of 120k pixels, while
keeping the aspect ratio unchanged. 

\paragraph{Local description.}
We compare two types of local descriptors that are aggregated in the same way. 
First, local SIFT descriptors~\cite{Low04} are extracted for each frame
on a dense grid~\cite{nowakECCV06}, every 4 pixels and at 5 scale
levels. 
We use dense sampling of local descriptors rather than interest points, as this increases the accuracy without impacting the storage size after aggregation.

Alternatively, dense trajectory descriptors (DT)~\cite{wang:hal-00873267} are extracted around dense points tracked during 15 frames.  We use the histogram of optical flow (HOF) descriptor for the trajectories, which was found to be most discriminant.
We refer to these features as DTHOF.
Note that for the last 15 frames of the video we do not extract additional 
trajectories, since tracks cannot be long enough. 

We compute the square-root of the local descriptor components and reduce the
descriptor to 32 dimensions with principal component analysis
(PCA)~\cite{AZ12,JBGJ12}. 

\paragraph{Descriptor aggregation.} The local
descriptors of a frame are encoded using MultiVLAD~\cite{JC12}, a variant
of the Fisher vector~\cite{sanchez13ijcv}. 
Two VLAD descriptors are obtained based on two  
different codebooks of size 128 and concatenated. 
Power normalization is applied to the vector, and it is reduced by PCA to
dimension $d$ (a parameter of our approach). 
The vector is normalized using the PCA covariance matrix and $\ell_2$-normalization.

\paragraph{Parameters of the frame descriptors.}
The parameters of the local descriptor extraction and VLAD aggregation were chosen on the
image retrieval benchmark Holidays~\cite{JDS08} to optimize the
trade-off between descriptor size, extraction speed and performance.
The Holidays dataset is an instance retrieval benchmark, 
which is closely related to the specific video event retrieval task addressed in this paper.

Because the descriptors are extracted from potentially large video
sequences, they need to be fast to compute and compact. 
Our implementation extracts frame descriptors based on SIFT in
real time (15~fps) on a single processor core, producing VLAD descriptors of size 512 or 1024.
For the dense-trajectory descriptor DTHOF, we used the default
parameters but only compute its histogram of flow (HOF)
descriptor. Various other settings and descriptors of the
dense-trajectory descriptor~\cite{wang:hal-00873267} were tested but
did not improve the results. 
Using trajectory descriptors is about ten times slower than
SIFT. Therefore, we perform most large-scale evaluations using SIFT
only. 

\subsection{Matching frame descriptors}
\label{sec:matching}

Our goal is to compare two sequences of 
frame descriptors $\qu=\left[\qu_1,\dots,\qu_m\right] \in \mathbb{R}^{d \times m}$ 
and $\b=\left[\b_1,\dots,\b_n\right] \in \mathbb{R}^{d \times n}$ corresponding to the query video and an item in the video database.

We first consider the metric 
\begin{equation}
s_{\delta}(\qu,\b)=\sum_{t = -\infty}^{\infty}\left\langle \qu_{t},\b_{t-\delta}\right\rangle ,
\label{eq:sdelta}
\end{equation}
where the vectors $\qu_t$ (resp., $\b_t$) are zero when $t<1$ and $t>m$ (resp., $t>n$). 
This is an extension of the \emph{correlation} used for pattern detection in scalar signals~\cite{Vijara05}.
The metric $s_\delta(\qu,\b)$ reaches a maximum in $\delta$ when the $\qu$ and $\b$ are aligned 
if the following assumptions are satisfied:
\medskip

{\noindent \it Assumption 1: There is no (or limited) temporal acceleration}. %
This hypothesis is assumed by the ``temporal Hough transform''~\cite{DJSP10} 
when only the shift parameter is estimated. 
\medskip

{\noindent \it Assumption 2:  The inner product is a good similarity measure between frame descriptors}. 
This is the case for Fisher and our MultiVlad descriptors (Section~\ref{sec:descriptors}), 
but less so for other type of descriptors, eg. bag-of-words, to be compared with more general kernels. 
\medskip

{\noindent \it Assumption 3: The sum of similarities between the frame descriptors at a constant time shift is a good indicator of the similarity between the sequences}. 
In practice, this assumption may not be satisfied. 
Videos are often self-similar in time. In this case, two long nearly static sequences may ---if their frame descriptors are somewhat similar---  accumulate to a stronger response than shorter sequences of truly temporally aligned frames.
Therefore, the similarity measure proposed in Equation~(\ref{eq:sdelta}) may favor an incorrect alignment. 
In our case, this problem will be addressed by a regularization of the metric in the Fourier domain
which also makes the metric invariant to video duration, see Section~\ref{sec:regmetric}. 
\medskip

We now present \emph{\ourmethod} (\OUR) to efficiently compute the
$s_{\delta}(\qu,\b)$ for all possible values of $\delta$ at once.
It  relies on Fourier-domain processing. Section \ref{sec:regmetric} presents the regularization techniques that address the limitations mentioned in Assumption~3. 

\subsection{Circulant encoding of vector sequences}
\label{sec:circ}

\newcommand{\mydot}{\centerdot}
\newcommand{\conjugate}[1]{\overline{#1}}

Equation~\ref{eq:sdelta} can be decomposed along the
dimensions of the descriptor. Introducing the per-row notations
 $\qu=[\qu_{\mydot 1}^\top, \dots, \qu_{\mydot d}^\top]^\top$
and $\b=[\b_{\mydot 1}^\top, \dots, \b_{\mydot d}^\top]^\top$, the scores for all possible values of $\delta$ is given by
\begin{equation}
\s(\qu,\b) = [\dots, \s_0(\qu, \b), \s_1(\qu, \b), \dots] = \sum_{i=1}^d \qu_{\mydot i} \otimes \b_{\mydot i},
\label{eq:conv}
\end{equation}
where $\otimes$ is the convolution operator. 
Assuming sequences of equal lengths ($n=m$), $\s(\qu,\b)$ is computed in the Fourier domain~\cite{Vijara05}. 
Denoting by $\mathcal{F}$ the 1D-discrete Fourier transform and  $\mathcal{F}^{-1}$ its inverse, the convolution theorem states that:%
\begin{equation}
\s(\qu,\b) 
= \sum_{i=1}^d \mathcal{F}^{-1} \left( \conjugate{\mathcal{F}(\qu_{\mydot i})} \odot \mathcal{F}(\b_{\mydot i}) \right)  
\label{eq:fouriercomp} 
\end{equation}
where $\odot$ is the element-wise multiplication of two vectors and $\conjugate{\mathcal{F}(\qu_{\mydot i})}$ is the complex conjugate of $\mathcal{F}(\qu_{\mydot i})$. 
Using row vector notations $\ftqu_i = \mathcal{F}(\qu_{\mydot i})\in\mathbb{C}^m$ and 
$\ftb_i = \mathcal{F}(\b_{\mydot i})\in\mathbb{C}^n$, the linearity of the Fourier operator gives: 
\begin{equation}
\s(\qu,\b)= \mathcal{F}^{-1} \left( \sum_{i=1}^d   \conjugate{\ftqu_i} \odot \ftb_i\right), 
\label{eq:fouriercomp2}
\end{equation}
which is more efficient to compute than Equation~\ref{eq:fouriercomp} 
because it requires a single inverse FFT instead of $d$, while performing the same 
number of component-wise multiplications.

In practice, we rely on the Fast Fourier Transform (FFT) and its inverse, which are very efficient, especially for sequences whose length is a power of two. As a common practice, 
the descriptor sequences are padded with zeros to reach the next power of two~\cite{Vijara05}. 
Unless stated otherwise, we consider hereafter that the sequences have been preprocessed to have the same length $m=n=2^\ell.$ 

\subsection{Regularized comparison metric}
\label{sec:regmetric}

\begin{figure}
\hspace*{-0.7em}
\begin{tabular}{c}
\includegraphics[width=\linewidth]{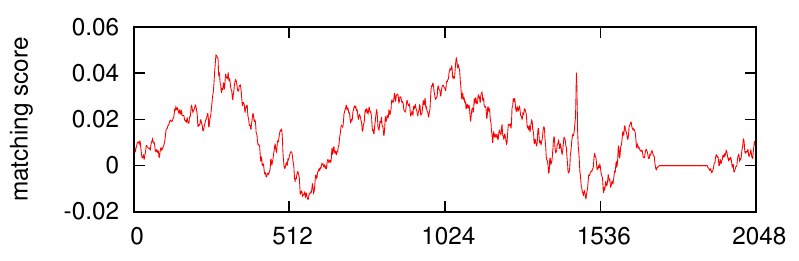}\\
(a) Raw matching score $\s(\qu,\b)$.\\
\includegraphics[width=\linewidth]{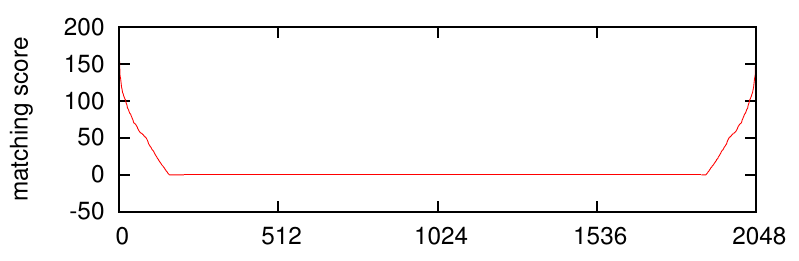}\\
(b) Raw self-matching score $\s(\qu,\qu)$ (zero-padded query).\\
\includegraphics[width=\linewidth]{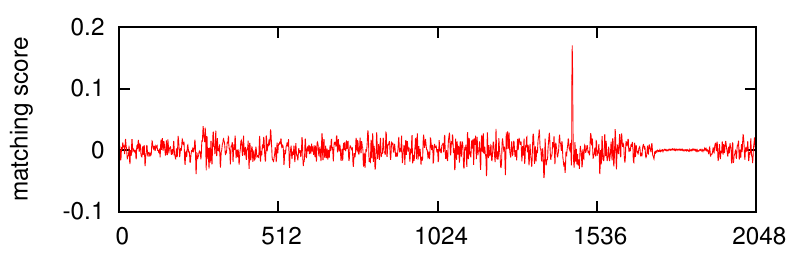}\\
(c) Regularized matching score $\s^\lambda(\qu,\b)$.\\
\end{tabular}
\caption{\label{fig:deltat}
  Matching score between a query $\qu$ and a database video $\b$ as a function of the time shift $\delta$. 
}
\end{figure}

As mentioned above, due to the temporal consistency and more generally the self-similarity
of frames in videos, the values of the score vector
$\s(\qu,\b)$ are noisy and its peak over $\delta$ is not precisely localized, as shown in Figure~\ref{fig:deltat}(a). This is obvious when comparing the query to itself: 
$s(\qu,\qu)$ consists of a relatively wide peak around $\delta=0$ whose
magnitude directly depends on the query duration, as shown in Figure~\ref{fig:deltat}(b).
Instead, one would ideally obtain a Dirac response, independent from the video content and duration, that is:
$s_{\delta}(\qu,\qu)=0$ for $\delta \neq 0$, and $s_{0}(\qu,\qu)=1$.
This behavior can be achieved through an additional filtering stage in the
Fourier domain. 
We use a set of filters $W=\{W_1,\dots,W_d\}$, $W_i \in \mathbb{R}^{n}$  to redefine the score vectors as
\begin{align}
\s^{W}(\qu,\b) & =  \mathcal{F}^{-1}\left(\sum_{i=1}^{d} \frac{\conjugate{\ftqu_i}\odot\ftb_i}{W_{i}}\right),\label{eq:ScoreWithFilter}
\end{align}
where the division is element-wise. 
We then aim to set the filters so that the score vector that compares the query with itself gives the desired Dirac response
\begin{align}
\s^{W}(\qu,\qu) & =  \mathcal{F}^{-1}\left(\sum_{i=1}^{d} \frac{\conjugate{\ftqu_i}\odot\ftqu_i}{W_{i}}\right) \nonumber \\
& = [1,0,\ldots,0] = \eone.\label{eq:dirac}
\end{align} 

For the sake of simplicity, we compute $W_i$ assuming that the contributions are shared equally across dimensions: 
\begin{equation}
\mathcal{F}^{-1}\left(\frac{\conjugate{\ftqu_i} \odot\ftqu_i}{W_{i}}\right) = \frac{1}{d}\eone\ \ \ \forall i =1\dots d
\end{equation}
Hence
\begin{equation}
\frac{\conjugate{\ftqu_i}\odot\ftqu_i}{W_{i}} = \frac{1}{d}\mathcal{F}\left(\eone\right)=\frac{1}{d}[1,1,\ldots1],
\label{equ:nonregfilter}
\end{equation}
The closed-form solution of each filter $W_i$ is therefore obtained in the Fourier domain as
\begin{equation}
W_{i}=d\ \conjugate{\ftqu_i}\odot\ftqu_i.
\label{equ:filterWi}
\end{equation}
The filters $W_i$ can be interpreted as a peak detectors in $\s(\qu,\b)$. 
In practice, its spectrum resembles that of a Laplacian filter.

One  drawback of this solution is that some elements of the denominator $W_i$ in Eqn.~\ref{eq:ScoreWithFilter} may 
be close to zero, which  magnifies the noise and de-stabilizes the solution. 
To tackle this issue, Bolme et al.~\cite{Bolme2010} average the filters
obtained from independent samples (dimensions in our case). 
It is indeed unlikely that high-dimensional filters obtained independently 
have near-zero values at identical positions.
In our case, we average the filters $W_i$, since they are independent due to 
the decorrelation property of the PCA \cite{Bis07}, see Section \ref{sec:descriptors}.

Unfortunately, averaging does not always suffice, as some videos 
contain only one shot composed of static frames: the components associated with high frequencies are near zero for all $W_i$. 
Therefore, we incorporate a regularization term into Equation~\ref{eq:dirac} and determine the $W_i$  minimizing
\begin{align}
\lambda\left\Vert \frac{1}{W_{i}}\right\Vert ^{2} 
+ \left\Vert \mathcal{F}^{-1}\left(\frac{\conjugate{\ftqu_i}\odot\ftqu_i}{W_{i}}\right)-\frac{1}{d}\eone\right\Vert ^{2}, 
\end{align}
where the regularization coefficient $\lambda$ ensures the stability of the filter. 
Notice that setting $\lambda=0$ amounts to solving Equation~\ref{equ:nonregfilter} and leads to the 
solution proposed in Equation~\ref{equ:filterWi}. A closed-form solution to this minimization problem
in the Fourier domain~\cite{HCMB12}, obtained by leveraging properties of circulant matrices,
consists of adding $\lambda$ to all the elements of the denominator in Equation~\ref{equ:filterWi}. This leads to a
regularized score vector between two video sequences $\qu$ and $\b$:
\begin{align}
\s^{\lambda}_\mathrm{reg}(\qu,\b) & =  \frac{1}{d} \mathcal{F}^{-1}\left(\sum_{i=1}^{d} 
\frac{
\conjugate{\ftqu_i} \odot\ftb_i}
{\conjugate{\ftqu_i}\odot\ftqu_i+\lambda}
\right). 
\end{align} 

Both regularization techniques, \ie, averaging the 
filters and using a regularization term, are complementary. 
We have empirically observed that combining them gives more stable results than 
using either one separately. This yields the following regularized matching score, 
illustrated in Figure~\ref{fig:deltat}(c):
\begin{align}
\s^{\lambda}(\qu,\b) & =  \mathcal{F}^{-1}\left(\sum_{i=1}^{d} 
\frac{
\conjugate{\ftqu_i} \odot\ftb_i}
{\ftqu_{\mathrm{den}}} 
\right), 
\label{equ:scorefinal}
\end{align} 
with 
\begin{align}
\ftqu_{\mathrm{den}} = {\sum_{j=1}^{d}{\conjugate{\ftqu_j}\odot\ftqu_j}+\lambda}.
\end{align} 
The choice of $\lambda$ is discussed in Section~\ref{sec:experiments}.

\subsection{Temporal alignment and boundary detection}
\label{sec:boundetect}

The strategy presented above produces a vector of matching scores, 
$\s^{\lambda}(\qu,\b)=[\dots,s^\lambda_\delta(\qu,\b),\dots]$, between two video sequences $\qu$ and $\b$, for all possible temporal offsets $\delta$. 
The optimal temporal alignment of the two videos is given by
\begin{equation}
\label{eq:delta}
\delta^*=\arg\max_{\delta \in {\mathbb Z}} s^\lambda_\delta(\qu,\b),
\end{equation}
 and $s^\lambda_{\delta^*}(\qu,\b)$ is their similarity score. 

In some applications such as video alignment (see Section~\ref{sec:experiments}), 
we also need the boundaries of the matching segments. For this purpose, the database descriptors are 
reconstructed in the temporal domain from $\mathcal{F}^{-1}(\b_{\mydot i})$. 
A frame-per-frame similarity is then computed with the estimated shift $\delta^*$ as:
\begin{equation}
S_t = \left\langle \qu_{t},\b_{t-\delta^*}\right\rangle.
\end{equation}
The matching sequence is defined as a set of contiguous $t$ for which the scores $S_t$ are high enough.
In practice, we use a threshold equal to half the peak score $\max_t S_t$.
This also refines the  matching score between videos, which is replaced by the sum of $S_t$ on the matching sequence.

Note that, unlike the computation of $\s^{\lambda}_{\delta^*}(\qu,\b)$, 
this processing requires $d$ distinct 1D inverse FFT, one per component. Yet, on large datasets 
this does not impact 
the overall efficiency, since it is only applied to a shortlist of database videos $\b$ with the highest pairwise scores $s^\lambda_{\delta^*}(\qu,\b)$.

%% file: indexing.tex
\section{Indexing strategy and complexity}
\label{sec:indexing}

This section discusses the steps  to efficiently 
encode the descriptors in the Fourier domain.
The goal is to implement the method presented in Section~\ref{sec:method} in an approximate manner. 
Beyond the complexity gain already obtained from our Fourier-domain processing, 
this considerably improves the efficiency of the method while reducing
its memory footprint by orders of magnitude.  
As shown later by our experiments (Section~\ref{sec:experiments}), this gain is achieved
without significantly impacting the retrieval quality.

\subsection{Frequency-domain representation}
\label{sec:frequencydomain}

A  video $\b$ of length $n$ is represented in the Fourier domain by a complex matrix 
$\ftb=[\ftb_1^\top,\dots,\ftb_d^\top]^\top=[\f_0,\dots,\f_{n-1}]$, with $\ftb \in \mathbb{C}^{d \times n}$. 
Our input descriptors are real-valued, so only half of the components %
are stored, as $\f_{n-i}$ is the complex conjugate of $\f_{i}$. 

\paragraph{Frequency pruning} is applied to reduce the video representation by keeping only a 
fraction $n' = \beta n \ll n$ of the low-frequency vectors $\f_i$, $i=0 \dots n'-1$ (in practice, 
$\beta$ is a negative power of 2). We keep a fraction rather than a fixed number of 
frequencies for all videos, as this would make the localization accuracy dependent on the sequence length.

\paragraph{Descriptor sizes.} 
 The comparison of a query $\qu$ video of length $m$ and a database video $\b$ of size $n$ in the Fourier domain requires them to have the same size. %
To handle the case $m \leq n$, we precompute a Fourier descriptor for different zero-padded versions of the query, 
\ie, for all sizes $2^\ell$ such that $m \leq 2^\ell \leq n_{\max}$, 
where $n_{\max}$ is the size of the longest database video.

We handle the case $m > n$ by noticing that the Fourier descriptor of the 
concatenation of a signal with itself is given by 
$\left[ \f_0, 0, \f_1, 0, \f_2, 0, \dots \right]$. Therefore, 
expanded versions of database descriptors can be generated on the fly and at no cost. 
However, this introduces an ambiguity on the alignment of the query and database videos: $\delta^*$ is determined modulo $n$, due to the repetition of the signal. In practice, the ambiguity is resolved when estimating the  matching segment's boundaries,  as described in Section \ref{sec:boundetect}.

\subsection{Complex PQ-codes and metric optimization} 

In order to further compress the descriptors and to efficiently compute
Equation~\ref{equ:scorefinal}, we propose two extensions of the product
quantization technique~\cite{JDS11}, a compression technique
that enables efficient compressed-domain comparison and search. 
The original technique proceeds as follows. 
A given database vector $y \in {\mathbb
  R}^{d}$ is split into $p$ sub-vectors $y_i$, $i=1\dots p$, of length $d/p$, where $d/p$ is an integer. %
  The sub-vectors are separately quantized using k-means
quantizers $q_i(.),\ i=1\dots p$. This produces a vector of indexes
$[q_1(y_1),\dots,q_p(y_p)]$. Typically, $q_i(y_i) \in [1,\dots,2^8]$. 

The comparison between a query descriptor $x$ and the database vectors
is performed in two stages. First, the squared distances between each
sub-vector $x_j$ and all the possible centroids are computed and stored
in a table $T=[t_{j,i}] \in \mathbb{R}^{p \times 256}$. This
step is independent of the database size. Second, the squared distance
between $x$ and $y$ is approximated using the quantized database vectors as 
\begin{equation}
d(x,y)^2 \approx \sum_{j=1}^p t_{j,q_j(y_j)},
\label{equ:pqestimation}
\end{equation}
which only requires $p$ look-ups %
and additions. 

We adapt this technique to our context in two ways. First, it is extended to complex vectors in a straightforward manner. 
We learn the k-means centroids for complex vectors by considering a $d$-dimensional complex vector to 
be a $2d$-dimensional real vector, and this for all the frequency vectors that we keep: 
${\mathbb C}^d \equiv {\mathbb R}^{2d}$ and $\f_j \equiv y_j$. At query time, the table $T$  stores complex values.

Computing the score of Equation~\ref{equ:scorefinal} requires to produce the $n'$ scalar products
between the corresponding columns of $\ftb$ and a matrix $\ftqu_\mathrm{reg}$, which represent a database video and the query video respectively, with 
\begin{equation}
\ftqu_\mathrm{reg}^\top = \left[
\left( \frac{\conjugate{\ftqu_1}}{\ftqu_{\mathrm{den}}} \right) ^\top 
\cdots
\left(\frac{\conjugate{\ftqu_d}}{\ftqu_{\mathrm{den}}} \right) ^\top
\right].
\end{equation}

Matrix $\ftb$ is approximated with a product quantizer, so for each
$d/p$-sized subvector of the $n'$ columns of $\ftqu_\mathrm{reg}$, we
precompute the scalar products with the 256 possible subvectors of
$\ftb$.

As a result, our table $T$ stores the partial sums for all possible centroids, 
including the processing associated with the regularization filter. 
As with the regular product quantization technique, a single comparison only requires 
$pn'$ look-ups and additions of complex numbers. The memory used for $T$ 
 ($2 \times 256 \times p \times n'$)  is a constant that does not depend on the database size. 

Interestingly, the product quantization vocabularies do not need to be learned on 
representative training data: they can be trained on random Gaussian vectors in 
$\mathbb{R}^{(2d/p)}$. This is due to the fact that the  PCA whitening applied to generate 
$\b_j$ and the Fourier transform applied on $\b_{\mydot i}$ normalize and decorrelate the 
signal spatially and temporally~\cite{Bis07,Mallat08}. 
We also L2-normalize each frequency vector $\f_i$ so that they have unit variance.
Empirically,  we verified that the resulting vectors are close to a Gaussian distribution.

\subsection{Summary of search procedure and complexity}

Each database video is processed offline as follows: 
\begin{enumerate}
\item The video is pre-processed, and each frame is described as a $d$-dimensional 
MultiVlad descriptor. %
\item This vector is padded with zeros to the next
power of two, and mapped to the Fourier domain using $d$ independent 1-dimensional FFTs.
\item High frequencies are pruned: Only $n' = \beta \times n$ frequency vectors are kept. 
After this step, the video is represented by $n' \times d$-dimensional complex vectors. %
\item These vectors are separately encoded with a complex product quantizer, producing 
a compressed representation of $p \times n'$ bytes for the whole video. %
\end{enumerate}

At query time, the submitted video is described in the same
manner, except that the query is not compressed with the product quantizer. Instead, it is used to pre-compute the look-up tables used in asymmetric PQ distance estimation, as discussed above. 
The complexity at query time depends on the number~$N$
of database videos, the size~$d$ of the frame
descriptor, and the video length, that we assume for readability to be constant ($n$ frames):
\begin{enumerate}
\item ${\O(d \times n \log n)}$ -- The query frame descriptors are mapped to the frequency domain by $d$ 1D FFTs of size $n$.
\item $\O(256 \times d \times n')$ -- The PQ table $T$ associated with the regularized query $\ftqu_\mathrm{reg}$ is pre-computed
\label{enu:pqtables}
\item ${\O(N \times p \times n')}$ -- Equation~\ref{equ:scorefinal} is evaluated for all database 
vectors using the approximation~of Equation~\ref{equ:pqestimation}, directly in the compressed domain using $n'\times p$ look-ups 
from $T$ and additions. This produces a $n'$-dimensional vector for each database video.  
\label{enu:pq}
\item ${\O(N \times n' \log n' )}$ -- This vector is mapped to the temporal domain using a single inverse FFT.
 Its maximum gives the temporal offset $\delta^*$ and the score $s^\lambda_{\delta^*}$.  
\label{enu:ifft}
\end{enumerate}

As described in Section~\ref{sec:frequencydomain}, the operations 1 and 2 are repeated for all sizes $n=2^\ell$ found in the dataset. 
This doubles the runtime of the operations applied to $n=n_\mathrm{max}$.
Only the steps~\ref{enu:pq} and~\ref{enu:ifft} depend on the database size. They dominate the complexity for large databases.

%% file: video_align.tex
\section{Global temporal video alignment}
\label{sec:videoalign}

In this section we produce a global temporal alignment of a collection of videos.
Aligning a set of videos requires to estimate the starting time of
each video on a shared time line, so that the temporal offsets estimated
between pairs of videos are best satisfied like in Figure~\ref{fig:AlignGraph}-right.  
The estimated common time line enables synchronized playback of videos captured from  different viewpoints, as depicted in Figure~\ref{fig:evveex}. 
We use the technique of Section~\ref{sec:boundetect} to match and estimate a temporal offset $\delta^*$ for all possible videos pairs.
Because of mismatches, edited input videos, \etc, the global alignment needs to be robust to outliers.

The temporal alignment is applied either to all result videos for a given query, or to a set of videos that are known a-priori to correspond to a single event. In Section~\ref{sec:RobustAlignment} we discuss how our approach can be applied to align single-shot videos (rushes), and Section \ref{sec:alignedited} extends the method to edited videos.

\subsection{Robust alignment from pairwise matches}
\label{sec:RobustAlignment}

We use the CTE method of  Section~\ref{sec:method} to find a collection of 
pairwise matches $(i,j)$   among the $N$ videos. Each match is characterized by $(\delta_{ij}, s_{ij})$, where $\delta_{ij}$ gives the estimated starting time of video $j$ in video $i$, and $s_{ij}$ is the confidence score $s^\lambda_{\delta^*}$  associated with the match. We retain matches only if $s_{ij}$ is above a threshold.
Aligning the videos consists in putting the videos on a common timeline by estimating the starting times $t_i$ for each video $i\in \{1,\dots,N\}$ such that for all matches $(i,j)$ we have 
$ t_j \approx t_i + \delta_{ij}$.

\begin{figure}
\includegraphics[width=\linewidth]{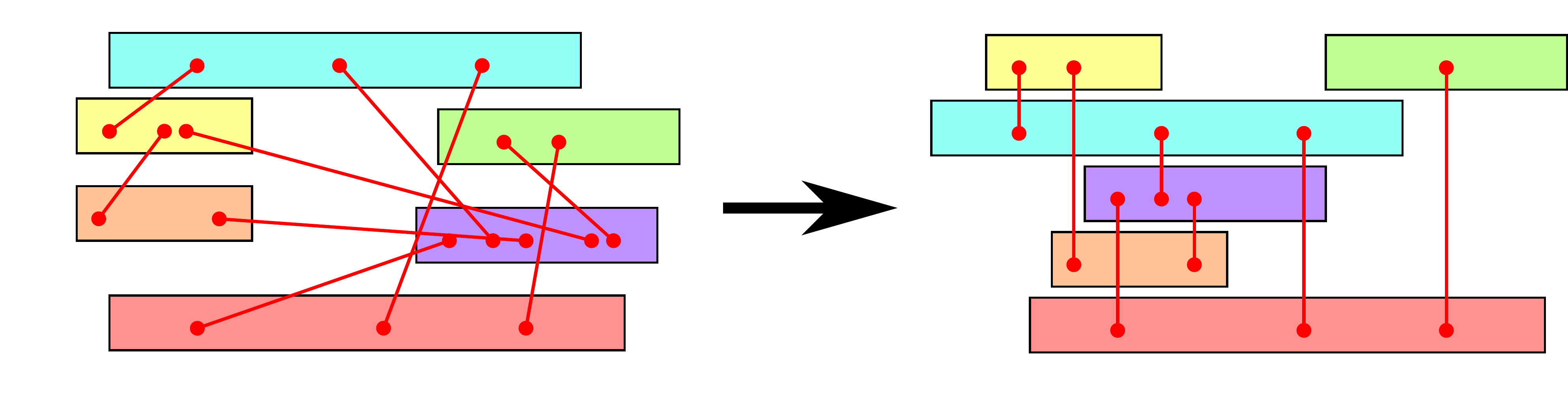}
\caption{\label{fig:AlignGraph} 
Illustration of the alignment process of multiple videos: some pairwise matches are removed to allow for a consistent global alignment.
}
\end{figure}

The input matches are noisy because the events may be similar but not the same (\eg, recurring similar motions observed in a sequences of a walking person), because the video content overlaps only in part of the videos (\eg, due to pan-tilt camera motions), or in case of edited videos where $\delta$ is valid only for part of the video. 
Therefore, in the alignment process we aim at finding a subset of matches that is as large as possible, 
while ensuring a consistent placement on a common timeline, up to a tolerance $\tau$.
More formally, given a set of $N$ videos and a set of $M$ matches $\mathcal{M} = \{(i_1,j_1), \dots, (i_M,j_M)\}$, the goal is to find a set $\mathcal{C} \subseteq \mathcal{M}$ that is as large as possible such that 
\begin{align}
\exists \{t_1, \dots, t_N\} \;\forall (i,j) \in \mathcal{C}:
 |t_i-t_j-\delta_{ij}| <\tau,
\end{align}
where $t_i$ is the starting point of video $i$ on the global timeline.
See Figure~\ref{fig:AlignGraph} for an illustration.

The pairwise  matches induce a graph over the videos.
Each node corresponds to a single video, and each match corresponds to an edge between two videos. 
The graph may contain several connected components if there are no edges between different groups of videos. 
For each connected component we produce a separate global alignment of the videos belonging to that component. 

\begin{figure}
\includegraphics[width=\linewidth]{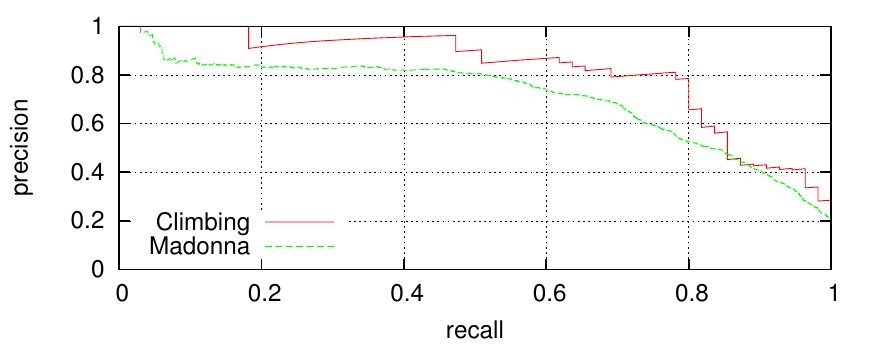}
\caption{\label{fig:prmatches}
	Precision-recall plot of the video anchor matches found by CTE, for SIFT descriptors of the \emph{Climbing} and \emph{Madonna} datasets. See Section~\ref{sec:dataset} for details on the datasets.
}
\end{figure}

We use an approximate greedy strategy to solve this problem. 
As shown in the precision-recall plot of Figure~\ref{fig:prmatches}, true matches often have higher scores than false matches. 
Therfore, the ordering of the match scores $s^\lambda_{\delta^*}$ tends to be reliable, so we first select a maximum spanning tree in the match graph based on the match scores.
Using this subset of matches and corresponding temporal offsets, we find an alignment that perfectly satisfies all the pairwise offsets, at least in the absence of additional constraints.
We then iteratively add  matches for which the temporal offset is consistent with the current alignment (up to a tolerance $\tau$),
and find the global alignment that minimizes the sum of squared errors $(t_i-t_j - \delta_{ij} )^2$ for all selected matches.
The procedure is detailed in Algorithm \ref{alg:alignment}.

Different connected components cannot be aligned automatically among each other due to the lack of matches. If they correspond to events that do overlap, manual alignment can be effective, since the number of connected components is typically much smaller than the number of input videos. Mismatches can also be corrected manually, see Section~\ref{sec:aligngui}.

\begin{algorithm}
\caption{\; Find globally consistent temporal  alignment }
\medskip
{\bf Input:}
\begin{itemize}
\item
set  of video matches $\mathcal{M} = \{(i_1,j_1), \dots, (i_M,j_M)\}$,
\item scores associated with the matches $s_1,  \dots, s_M$.
\item  tolerance threshold $\tau$, 
\end{itemize}
\medskip
{\bf Output:}
\begin{itemize}
\item  global temporal alignment $t_1,.., t_N,$ 
\item  subset of matches $\mathcal C \subset \mathcal M$  consistent with alignment
\end{itemize}
\bigskip
\begin{enumerate}
\item Compute maximum spanning tree (MST) based on the  scores $s_1,  \dots, s_M$.
\item Initialize $\mathcal C$ as set of edges in the MST.
\item Solve for alignment using subset of edges
\begin{align}
\min_{\{t_1,\dots, t_N\}} \sum_{(i,j) \in \mathcal{C}} (t_{i} - t_{j} - \delta_{ij}) ^ 2 
\label{eq:AlignError}
\end{align}
\item  
  Add matches $(i,j)\in \mathcal M$ with  $ |t_{i} - t_{j} - \delta_{ij}|<\tau$ to $\mathcal C$
\item  If  matches were added: go back to step 3,\\
else return the estimated $\{t_i\}$ and the consistent set of matches $\mathcal C$,
\end{enumerate}  
\label{alg:alignment}
\end{algorithm}

\subsection{Alignment of edited videos}
\label{sec:alignedited}

In order to align edited videos, we determine the temporal extent of the matches using the method of
Section~\ref{sec:boundetect}, which produces  hypotheses about matching sub-sequences; we
call these ``anchor segments''. 
The anchor segments replace videos as the basic units manipulated by Algorithm~\ref{alg:alignment}. 
Note that two anchor segments from the same video, but defined by independent matches, can temporally overlap.
Aligning the anchor segments instead of complete edited videos allows us to deal with cuts and non-monotonic timeline edits.

There are two types of links between anchor segments: matches produced by the CTE alignment matches (m-links), and links we add to ensure
that overlapping anchor segments in a same video are kept together (o-links). 
Contrary to the non-edited case, here we use the  scores computed from the matching sub-sequences for the m-links, as defined in Section~\ref{sec:boundetect}.
For the o-links we use the overlap length as score, and since the o-links are more reliable than the m-links, we add a constant that ensures that they are at least as high as  m-link scores.

On output, the algorithm provides one time offset $t_i$ per anchor segment.
In the parts of the video where anchor segments overlap, 
we select the offset of the anchor segment with highest
associated matching score, so that each video frame has a unique location in the
global timeline.

%% file: dataset.tex
\begin{figure*}
\includegraphics[width=\linewidth]{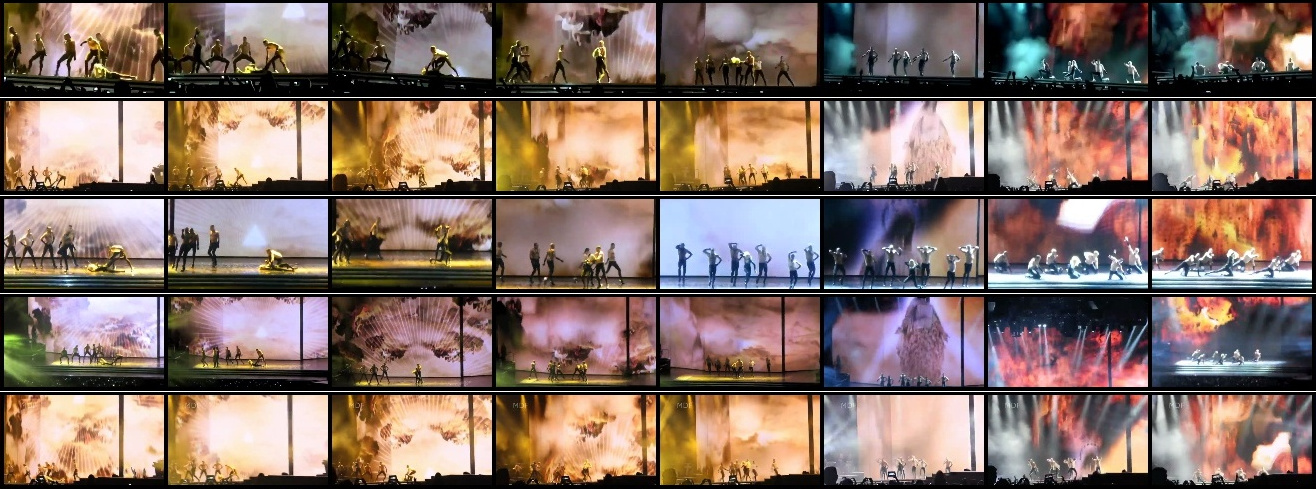}
\vspace{0mm}

\includegraphics[width=\linewidth]{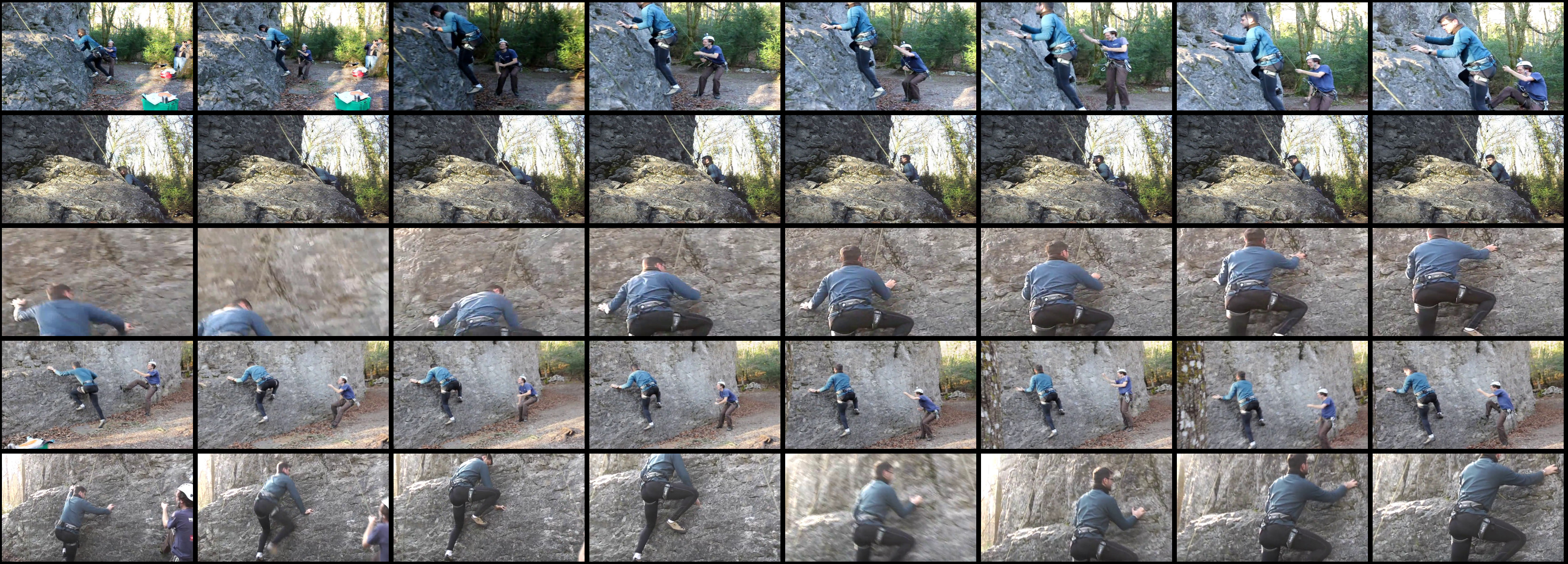}
\caption{\label{fig:evveex} 
Examples of correctly aligned clips of the \emph{Madonna} (top) and \emph{Climbing} (bottom) datasets. Each row is a different video, and each column corresponds to temporally aligned frames  (2 frames per second are represented).
}
\end{figure*}

\section{Datasets for video retrieval and alignment}
\label{sec:dataset}

In this section, we present the datasets used to evaluate our 
methods for large-scale video retrieval and for temporal video alignment. 
First, we describe the \ccweb and \trecvid 2008 copy detection datasets for large-scale video retrieval. 
Then we introduce the \emph{Climbing} and \emph{Madonna} datasets to evaluate 
the precision of temporal video alignment. These two datasets are publicly 
available for download. Table~\ref{tab:datastats} sums up statistical information on 
all the datasets used in our experimental evaluation.

\subsection{Content-based copy detection}

We evaluate large-scale video retrieval (\ie copy detection) on two public benchmarks, 
the \ccweb dataset~\cite{singinghippo}
and the \trecvid 2008 content based copy detection dataset\break 
(CCD)~\cite{SOK06}. 

\ccweb 
contains 24 query videos, mostly focusing on near-duplicate
detection. The transformed versions in the database correspond to user
re-posts on video sharing sites.  Large-scale performance is 
evaluated on \ccwebdist obtained by adding 100,000 video distractors taken from the
EVVE dataset~\cite{CTEPaper}. Performance is reported as the mean average precision (mAP) over all queries.

The 2008 campaign of the \trecvid CCD task is the last 
for which video-only results were evaluated. We present results on the %
camcording subtask, which is most relevant to our context of
event retrieval in the presence of significant viewpoint changes. We
report results with the official normalized detection cost rate (NDCR) measure. The NDCR is a weighted 
combination of the costs of 
false alarms and missed detections, so lower values are better.

\subsection{Video alignment: Climbing and Madonna}

We introduce two challenging datasets to evaluate our alignment method in
realistic conditions.\footnote{They are available at \url{http://pascal.inrialpes.fr/data/evve/index_align.html}.} Figure~\ref{fig:evveex} shows example frames for videos of the two datasets.

\paragraph{Climbing.} The dataset is composed
of video clips shot during an outdoor climbing session. The 
climbers as well as the people filming were all members of 
our lab. 

We used eleven different cameras of several types, including digital cameras, mobile phones, camcorders, head-mounted wide-angle cameras.  
The people filming were instructed to record sequences from several locations  with
different zooms, and to take relatively short clips ranging from 5~seconds
to 5 minutes. The raw video rushes from the cameras was used, \ie, each clip represents a single
shot. At each moment, there was at least one camera recording. 

The 89 videos were manually aligned to produce the ground-truth. The
precision of this ground-truth alignment is reliable to at least 0.5 second. Since each
clip represents a single shot,  there is a single time shift to estimate between each pair of clips.

Furthermore, there is additional metadata associated with the clips: 
the source camera and the ordering of the clips given by the clip's 
filename. This information is useful because
a camera cannot shoot two clips at a time, which could be used as a constraint on the alignment. Such constraints can be taken into account in the alignment algorithm in the form of linear inequality constraints. 
In the case where the relative times of the videos are known, it suffices to estimate a single starting time, and the videos can be considered as a single one for the purpose of temporal alignment. 
In our experiments we do not exploit this metadata, however, in order to simulate more challenging conditions.

From the ground-truth we can derive all pairs of videos that temporally overlap, and their respective temporal offsets.
To evaluate an alignment, we count the number of  pairwise overlaps in the ground-truth for which the correct temporal offset is recovered by our algorithm, within a tolerance of~0.5~s. This produces the \emph{pairwise alignment score} (PAS).
We compare this number of correct overlaps to the total number of overlaps in the ground-truth.

We always set the tolerance parameter $\tau$ of the alignment algorithm to the same value (of 0.5 s.) that is used in the evaluation. This is a natural choice, since setting the tolerance in the alignment algorithm larger (smaller) than the tolerance in the evaluation would lead to many false positive (negative) alignments.

\paragraph{Madonna.}  
This dataset is composed of videos from the\break
\emph{Madonna in Rome} event defined in the publicly available EVVE dataset~\cite{CTEPaper}. 
It shows one particular Madonna concert, that was filmed by dozens of people who posted their clips on
YouTube, often after editing them. There is a lot of stage action with distinctive motion, and recognition is facilitated by a large video screen on the side of the stage playing along with the concert.

The videos are shot with low-quality devices  such as mobile phones in
difficult shooting conditions: shaking cameras, large distance from the
scene, poor lighting and image quality.

Since the videos are edited, we annotated ground-truth segments on each video. A segment can cover shot boundaries, as long as the real time of the end of a shot coincides with the beginning of the next one. 
Then we annotated the ground-truth alignment on these ground-truth segments.

\begin{figure}
\includegraphics[width=\linewidth]{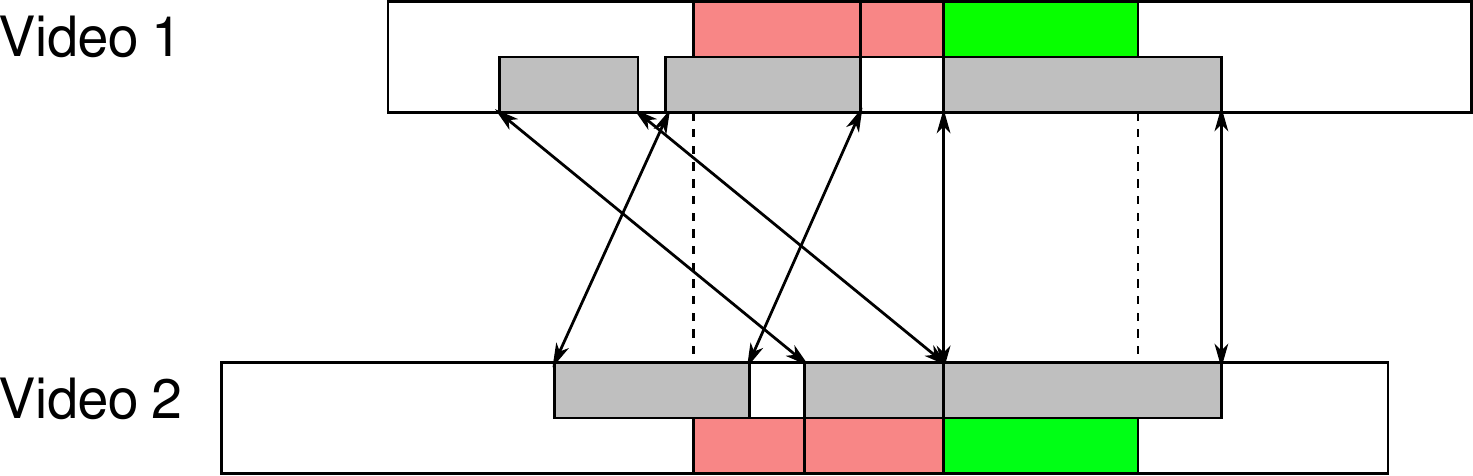}
\caption{\label{fig:eval}
A  match between ground-truth segments of two videos (combined red and green area), aligned horizontally. Anchor segments that are overlapping in the alignment output are indicated in gray and with arrows. The sub-segments of the ground-truth segment are counted as positive for the PAS evaluation are in green, the negative ones in red.
}
\end{figure}

\paragraph{Evaluation for edited videos.} 
As discussed in Subsection~\ref{sec:alignedited}, the alignment is defined
on anchor segments. Therefore, the estimated time shift is not uniform
over the intersecting range of ground-truth segments: on the example
of Figure~\ref{fig:eval}, considering the segment of Video~1, 
one part of the range has no
matching section at all on Video~2, another one has, but with an
incorrect time shift, and the last match is aligned correctly. 

In this case, we extend the PAS score by computing only the fraction
of the range that is aligned correctly, up to a tolerance of 0.5 s. The fractions are summed up
for all pairs of overlapping ground-truth segments, producing a
non-integer PAS.

%% file: experiments.tex
\section{Experimental results}
\label{sec:experiments}

In this section we evaluate our approach,  both for video copy detection 
and alignment of multiple videos belonging to the same event.
To compare the contributions of the
frame descriptors and of the temporal matching,  we introduce an
additional descriptor obtained by averaging the per-frame descriptors of 
\sect{descriptors} over the entire video. This 
descriptor that no longer carries temporal information is compared using the dot product and  denoted by
\ourbaseline (\BSL). 

\begin{table}
\centering

\begin{center}(a) Retrieval datasets\end{center}

\begin{tabular}{|l|c|ccc|}
\hline
dataset        & query  & \multicolumn{3}{c|}{database} \\
               & videos & videos & hours & frames \\
\hline
\ccweb & 24     & 13129  & 551   & 29.7M \\
\ccweb + 100k & 24     & 113129  & 5921   & 320M \\
\trecvid CCD 08 & 2010   & 438    & 208   & 11.2M \\
\hline
\end{tabular}

\begin{center}(b) Alignment datasets\end{center}

\begin{tabular}{|l|cccc|}
\hline
dataset        & \# videos  & \# g.t.\ segments   & \# groups    & length \\
\hline
Juggler     & 6 & N/A & 1 & 14m18s \\
Climbing    & 89 & N/A & 1 & 6h21m \\
Madonna     & 163 & 230 & 29 & 10h12m\\
\hline
\end{tabular}

\caption{\label{tab:datastats}
  Statistics on the different datasets used in this paper. 
}
\end{table}

\subsection{Video copy detection}

This task is evaluated on the \ccweb dataset~\cite{singinghippo}
and the \trecvid 2008 content based copy detection dataset\break
(CCD)~\cite{SOK06}.

\begin{figure}
\includegraphics[width=0.48\linewidth]{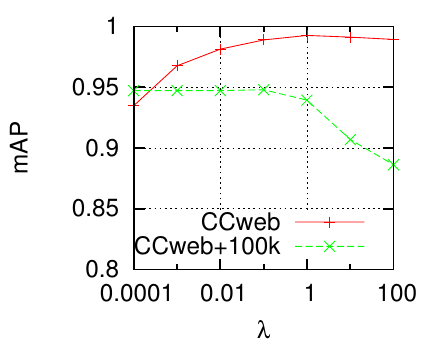}
\hspace{2mm}%
\includegraphics[width=0.48\linewidth]{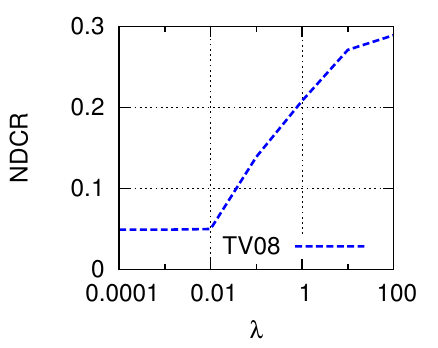}
\caption{\label{fig:lambda}Impact of the regularization parameter $\lambda$ on the performance.}
\end{figure}

\newcommand{\nf}[2]{#1/#2}

\def\mysep{{\hspace{10pt}}}
\begin{table}
\centering
{\small 
\begin{tabular}{|l@{\mysep}|@{\mysep}c@{\mysep}|@{\mysep}c|rrr|}
\hline
method      &  $p$ & $\beta$   & mAP   & memory    & search \\
            &    &           &         & usage     &  time \\
\hline
\multicolumn{6}{|c|}{\ccweb} \\
\hline
\multicolumn{3}{|l|}{HIRACH~\cite{singinghippo}} 
                              & 0.952   & -         & -     \\
\multicolumn{3}{|l|}{MFH~\cite{Song2011}}
                              & 0.954   & 0.5 MB    & -     \\

\BSL & no  & -         & 0.971   & 26.9 MB   & 1.5 ms  \\

\BSL & 64 & -         & 0.969   &  0.8 MB    & 0.7 ms   \\

\BSL & 16 & -         & 0.962   & 0.2 MB    & 0.5 ms \\

\OUR & no & \nf{1}{64} & \textbf{0.996}    & 2,600 MB    & 66.1 ms   \\

\OUR & no & \nf{1}{512}& 0.995   & 304 MB  &  4.8 ms  \\

\OUR & 64& \nf{1}{512} & 0.995   & 5.6 MB    &  1.0 ms \\

\OUR & 16& \nf{1}{512} & 0.994   & 1.4 MB    &  0.5 ms \\

\OUR & 16& \nf{1}{1024} & 0.990   & 0.7 MB    &  0.5 ms \\

\hline
\multicolumn{6}{|c|}{\ccweb + 100,000 distractors} \\
\hline
\multicolumn{3}{|l|}{MFH~\cite{Song2011}}
           & 0.866       & 5.3 MB    & 533 ms \\

\BSL & 16 & - &  0.887  & 1.8 MB    & 23 ms  \\

\OUR   & 16 & \nf{1}{1024} &  \textbf{0.960}  & 9.6 MB    &  75 ms \\ %

\hline
\end{tabular}}
\smallskip
\caption{\label{tab:ccdperf}
  Results for video copy detection on the \ccweb dataset in mAP (higher is better).
}
\end{table}

\def\mysep{{\hspace{10pt}}}
\begin{table}
\centering
{\small 
\begin{tabular}{|l@{\mysep}|@{\mysep}c@{\mysep}|@{\mysep}c|rrr|}
\hline
method      &  $p$ & $\beta$   & NDCR   & memory    & search \\
            &    &           &         & usage     &  time \\
\hline
\multicolumn{3}{|l|}{Best official result}
                          &  0.079      & 10,000 MB  & 16 min \\
\multicolumn{3}{|l|}{Douze et al.~\cite{DJSP10}}
                          &  0.224      & 300 MB     & 191 s \\

\BSL & no & - &  0.967     & 0.9 MB     & 4 ms \\

\OUR  & no  & \nf{1}{8}   & \textbf{0.049}      & 8,600 MB &  9.4 s \\

\OUR  & no  & \nf{1}{32}  & 0.077      & 2,150 MB & 2.2 s  \\

\OUR  & 64 & \nf{1}{8}   & \textbf{0.049}      & 134 MB  &  8.9 s  \\

\hline
\end{tabular}}
\smallskip
\caption{\label{tab:TV08perf}
  Results for video copy detection on the \trecvid dataset in terms of NDCR (lower is better).  
}
\end{table}

\paragraph{Compression parameters.} The spatial and temporal
compression is parametrized by the dimensionality~$d$ after
PCA, the number~$p$ of PQ sub-quantizers and the frame description
rate~$\beta$, which defines the ratio between the number of frequency
vectors and the number of video frames. 
As a general observation across all datasets and experiments, we notice that
higher values of~$d$ yield better performance, for all values
of~$p$. Yet $d$ should be kept reasonably small to avoid increasing the storage for the dataset 
(the memory usage we report is the size of the compressed dataset in RAM). 
We thus fix the PCA output dimension to $d=512$ in all our experiments 
and vary the number of sub-quantizers $p$ and the rate~$\beta$. 

\paragraph{Impact of the regularization parameter.} 
The choice of $\lambda$ depends on the task and the evaluation metric.  For near-duplicate
video retrieval, Figure~\ref{fig:lambda} shows that intermediate values of
$\lambda$ yield the best performance. In contrast, we observe that small values of
$\lambda$ produce the best NDCR performance for the \trecvid copy
detection task. This is probably due to the fact that the NDCR measure
strongly favors precision over recall,  whereas any matching tolerance
obtained by a larger $\lambda$ also produces more false
positives.   In all our experiments, we set $\lambda$=$0.1$ for the
near-duplicate detection task, and $\lambda$=$0.001$ for
the TV08 benchmark. 

\paragraph{Comparison with the state of the art.} 
Tables~\ref{tab:ccdperf} and \ref{tab:TV08perf} report our results  for near-duplicate and copy-detection for different compression trade-offs, and compares our results to the state of the art. 
  Search times are given for one core and are averaged across queries. See Section~\ref{sec:indexing} for details on the parameters $p$ and $\beta$.

On \ccweb, both the temporal and non-temporal versions of our method 
outperform the state of the art for comparable memory footprints. 
The good results of \BSL establishes  the quality of the image descriptors.
\OUR compresses the vector sequence by a factor 1024 along the temporal axis and by a factor 128 in the 
visual axis, which amounts to storing \emph{4 bits per second of video}. 
The results for the large-scale version of the dataset are not strictly comparable with those of the original paper~\cite{Song2011} 
because the distractor videos are different 
(they do not provide theirs).

On the \trecvid 2008 dataset, our approach is significantly more accurate than that of Douze \etal~\cite{DJSP10}, while being faster and using less than half the memory when using PQ compression. 
\BSL cannot be realistically evaluated on this dataset because it can not output temporal boundaries
for video matches. To compute its NDCR score, we disregard the boundaries, which are normally used 
to assess the correct localization of the matching segment within a video clip. %
Despite this advantage, \BSL performs poorly (NDCR close to 1), due to the small overlap between queries and 
database videos (typically 1\%), which dilutes the matching segment in the video descriptor. 

The results of CTE depend on the length of the subsequence shared by the query and retrieved videos. On Trecvid 2008, pairs with subsequences shorter than 5s are found with 62\% accuracy, subsequences between 5s and 10s with 80\% accuracy and 
longer subsequences with 93\% accuracy.

\paragraph{Timings.} Even for the largest dataset,  \ccweb with 100k distractors, 
the bottleneck remains the MultiVlad descriptor extraction, which is performed approximately in real-time 
on one processor core (\ie 1-2 minutes per query on \trecvid and \ccweb). 
Table~\ref{tab:ccdperf} shows that 
the search itself takes 23\,ms and 75\,ms on average for \BSL and \OUR, respectively, which is orders of magnitude faster than other methods with comparable accuracies. 

\subsection{Pairwise temporal alignment}

We first  evaluate the precision of our  pairwise alignment on relatively simple 
video sequences with a 
very precise ground-truth alignment. We chose the ``Juggler'' sequences from~\cite{pollefeyssiggraph},
that  are recorded from six
different camera viewpoints. They represent a dynamic event filmed in a street by
several people, with slowly moving cameras and high-resolution images. 
The main challenge is the large change in viewpoint between individual sequences. 
To simulate our use case, 
we resize and compress the videos to ``YouTube'' 360p format, reducing
each file from~1~GB to 15~MB.  
As the original videos span the same time range, and hence do not need alignment, 
we extract 5 random sub-videos of 15, 30, 60 and 120 seconds from each video,
and measure the alignment accuracy of these \wrt the other original videos.  
We measure the fraction of correctly aligned videos, with an alignment tolerance of 0.2 seconds.

\begin{figure}
\begin{center}
\includegraphics[width=\linewidth]{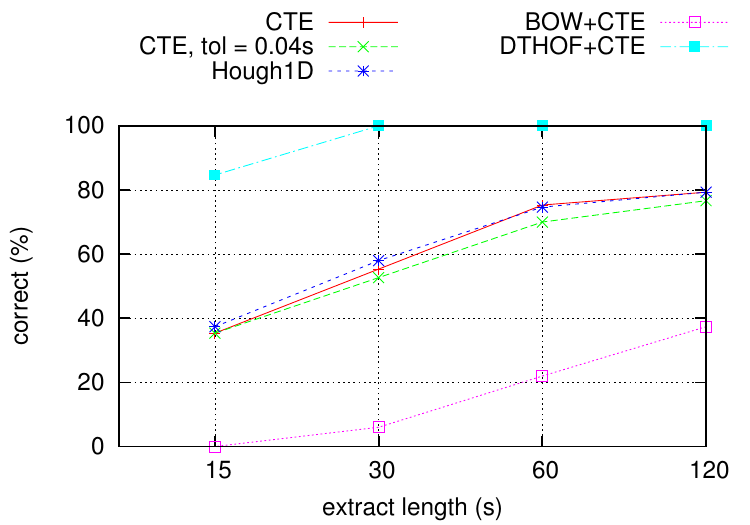}
\end{center}
\caption{\label{fig:alignerr} Fraction of correctly aligned video pairs from the ``Juggler'' sequence. The default alignment tolerance is 0.2~seconds, and  the default CTE setting uses SIFT descriptors with MultiVlad encoding. 
}
\end{figure}

Results in Figure~\ref{fig:alignerr} show  that pure content-based temporal alignment is possible 
despite strong viewpoint\break changes and the use of a global frame descriptor. 
Longer extracts help the alignment, and there is little difference between the alignment performance with a tolerance of 0.2~s and 0.04~s.  This shows that if the alignment is correct, it is usually
accurate to the frame. 
Furthermore, the DTHOF descriptors perform much better than SIFT, 
producing perfect alignments for all sequences when they are longer than 30\,s. 
This is expected as the dataset consists of a juggling performance for which motion is essentially discriminative.

We compare CTE with an alignment method based on the Hough Transform. 
Figure~\ref{fig:alignerr} show that this results in about the same precision as with CTE. 
Nevertheless, it is much slower: aligning  the 20~Juggler videos takes 745~s with Hough transform instead of 48~s for CTE. 

To evaluate the impact of our local descriptor aggregation strategy, 
we tried to replace the MultiVlad frame descriptor with bag of words (BOW) aggregation.
We used 1000 words, to match the MultiVlad descriptor size. 
similarly to an observation of~\cite{JC12}, the performance of BOW encoding is low.
This demonstrates that having a frame descriptor both invariant and discriminative is crucial.

\begin{table*}
\begin{center}
\begin{tabular}{|l||ccc|ccc|}
\hline
 &  \multicolumn{3}{c|}{Climbing} & \multicolumn{3}{c|}{Madonna} \\
 & SIFT & DTHOF & SIFT + DTHOF & SIFT & DTHOF & SIFT+DTHOF\\
\hline\hline
\# input anchors                & \multicolumn{3}{c|}{89}  & 11120 & 10906 & 10758\\
\# input matches                & 788  & 842 & 713 & 83691 (o) + 5726 (m) & 95629 (o) + 5695 (m) & 86104 (o) + 5593 (m)\\
\# correct input matches        & 55   & 54  & 71  & 73263 (o) + 1495 (m) & 86379 (o) + 1414 (m) & 76274 (o) + 1518 (m)\\
\hline
\# matches retained		        & 104  & 115 & 117 & 83691 (o) + 1711 (m) & 95629 (o) + 1923 (m) & 86104 (o) + 1922 (m)\\
\# correct matches retained     & 50   & 50  & 63  & 73263 (o) + 1142 (m) & 86379 (o) + 1202 (m) & 76274 (o) + 1245 (m)\\
\hline
PAS for ground-truth            &  \multicolumn{3}{c|}{470}  & \multicolumn{3}{c|}{1153} \\
PAS for alignment               & 109.8  & 80.7 & 132.4 & 632.2 & 518.0 & 656.3\\
\hline
\end{tabular}
\end{center}
\caption{\label{tab:AlignResults}
Alignment evaluation on the \emph{Climbing} and \emph{Madonna} datasets. The five first lines report statistics about the 
anchors (video fragments) input to the alignment algorithm, and their matches. The resulting Pairwise Alignment Score (PAS) is at the bottom of the table.	
}
\end{table*}

\subsection{Global alignment} 

In the following, we evaluate our pairwise and global matching methods
on the \emph{Climbing} and \emph{Madonna} datasets. 
Since the datasets are small, it is possible to use high-quality settings to compute the pairwise video matches: 1024-dimensional frame descriptors, no PQ
compression and $\beta = 1/4$. We also set $\lambda=10^{-2}$, since it is closest to a copy detection task.
Table \ref{tab:AlignResults} presents the results of the alignment evaluation on the \emph{Climbing} and \emph{Madonna} datasets.

The first three rows show the number of input anchors and matches produced by pairwise CTE matching, as well as the number of matches that are consistent with the ground-truth. 
For the \emph{Madonna} dataset, we separate the number of regular matches (m-links) and the anchor overlaps (o-links), see Section~\ref{sec:alignedited}.
We observe that the pairwise matching is much harder for the \emph{Climbing} than for the \emph{Madonna} dataset. 
For \emph{Madonna}, about 90~\% of the overlap links are correct (incorrect ones are due to undetected shot boundaries), while about 25\% of the between-video matches are correct, depending on the input descriptor. 
The reason for this discrepancy is that for  \emph{Madonna} there are more distinctive visual features, especially due to the large screen display on the back of the stage that shows fast moving videos. For \emph{Climbing} there are more repeated features of rock and foliage textures, larger baseline between viewpoints and limited action,  see Figure \ref{fig:evveex}. 
Rows four and five show the number of consistent matches in $\mathcal C$ that was retained  to produce the alignment, and the number of retained matches consistent with the ground-truth.
For \emph{Climbing}, only about 100 matches are retained, which contain 50 (of the 54 or 55) matches that are consistent with the ground-truth. 
For \emph{Madonna}, all overlap links are retained and about 1800 real matches, 60~\% of these correct.

Finally, the last two rows show the pair wise alignment score (PAS) obtained for the ground-truth, 
and using the produced alignment. 
For \emph{Madonna} the alignment obtains a PAS of 54\% relative to the ground-truth, 
while for \emph{Climbing} this is only 23\%.
This is in-line with the much higher fraction of correct retained matches for the \emph{Madonna} dataset. 

The results with the DTHOF descriptor are significantly lower. A possible explanation is that, compared to the Juggler dataset, the matching must not only distinguish subtle motions (for which DTHOF performs well), but also identify different environments and background patterns (where SIFT is more appropriate). 

It is possible to combine both descriptors by concatenating them frame per frame. Since the descriptors are normalized, 
concatenating them produces a vector of norm 2, that can be indexed with CTE. The combination SIFT+ DTHOF significantly improves the PAS, which shows that they are complementary.

\subsection{User evaluation of video alignment}
\label{sec:aligngui}

\begin{figure}
\includegraphics[width=\linewidth]{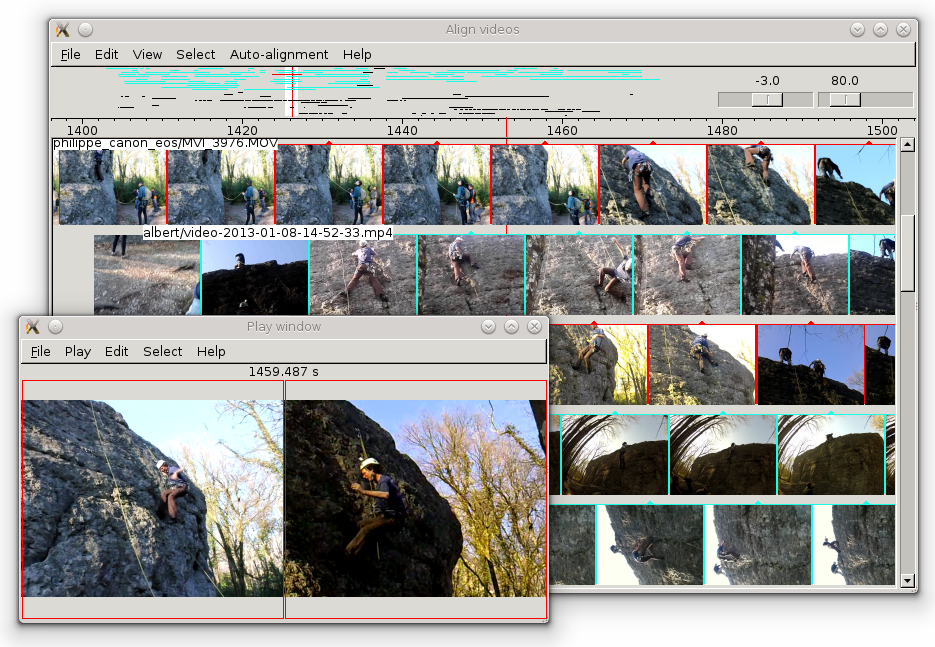}
\caption{\label{fig:aligngui} 
  \aligngui interface showing two aligned clips selected by the user in red, and playing them synchronously  for verification. The horizontal widget on top shows an overview of all aligned clips. 
}
\end{figure}

We also performed a user study to evaluate the quality of our 
video alignment. For this evaluation we developed a tool, \aligngui\footnote{Publicly available at~\url{http://pascal.inrialpes.fr/data/evve/index_align.html}.},
which allows to interactively align videos.
The user manipulates  anchor segments (potentially matching sub-sequences detected by the algorithm) extracted from several videos. 
In the interface, each video is displayed on a single line, and the anchor segments as a horizontal strip of frames,  see Figure~\ref{fig:aligngui}. 
Different from standard video editing tools, \aligngui can display several videos simultaneously, and plays them synchronized with the current alignment hypothesis.

Anchor segments can be aligned manually or interactively. 
In the manual mode, the user drags anchor segments on the timeline until they are aligned.
In the interactive mode, the user selects an anchor segment, and \aligngui uses
the matching method of Section~\ref{sec:method} to generate a list 
of alignment hypotheses for the other anchor segments. 
The user then cycles through the hypotheses until the correct alignment is found. 

Given four videos of the \emph{Climbing} data, we evaluate how long it
takes a user to align them with 48 already aligned clips.
We compare two operating modes.
In the manual mode, comparable to the workflow of conventional video editing tools,
the user performs all alignments by hand.
In the interactive mode the system suggests alignments to the user based on the pairwise matches, starting from the most confident one.
The user then cycles between the different propositions, and validates the one that seems correct to him.
Three users participated in this study: A is very familiar with both the data and software, B is less trained with the software but also knows  the video collection, and C was not familiar with either.  
The results are given in Table \ref{tab:gui}.

\begin{table}

\begin{center}
\begin{tabular}{|l|ccc|}
\hline
             & A     & B     & C \\
\hline
Manual & 11m 52s & 33m 18s   & 216m 57s\\
Interactive  & 1m 19s  & 1m 05s    &     1m 37s\\
\hline
\end{tabular}
\end{center}
\caption{Timing of three users (A,B, and C), performing manual or interactive video alignment. }
\label{tab:gui}
\end{table}

We observe that a manual alignment is at least ten times slower than an
interactive one, and even slower if the editor is not familiar with the software or the video collection. 
In both modes, the user has to play the videos synchronously to confirm each considered alignment. 
In the manual mode, however, the user must additionally first determine sub-sequences to match from a given video, search for potentially matching videos, and determine a pairwise temporal alignment. 
For user C that is unfamiliar with the video collection, the manual alignment becomes prohibitively slow.

%% file: conclusion.tex
\section{Conclusion}
\label{sec:conclusion}

The circulant temporal encoding technique proposed in this paper exploits 
the efficiency of Fourier domain processing to circumvent the usual frame-to-frame matching 
employed by most video matching techniques. This offers a 
very significant increase in efficiency, which is further magnified by the 
variant of product quantization that we introduce for complex-valued vectors. 
As a result, the matching time itself is faster than the query video duration by a factor 10 to 1000
(depending on the required matching accuracy), even for a collection of hundred thousands video clips. 
The bottleneck remains the (real-time) extraction of the frame descriptors associated with the video query, 
which suggests that our method is scalable to even  larger video collections, thanks to the 
low memory footprint of our compressed descriptors. 

Our approach is shown effective in the traditional copy detection scenario as well as in a more challenging 
 case,  when the same event is filmed from different viewpoints. %
Additionally, 
our approach produces an estimate of the relative time difference between two videos. 
Another contribution is our algorithm that exploits this information to align 
a set of matched videos on a global timeline. This allows us, in particular, to produce a synchronous playback of 
videos associated with a given event. 

To evaluate a global temporal alignment, we introduced the \emph{Climbing} and \emph{Madonna} datasets, for which we provide ground-truth temporal alignments annotated manually, and the frame descriptors that can be used to evaluate the alignment algorithm separately. These datasets are challenging, as the videos are self-similar on both short and long time scales. This corresponds to a realistic video-editing scenario. 

\section*{Acknowledgements}

We are grateful to the team members who participated as cameramen or actors in the shooting of the \emph{Climbing} dataset.
This work was supported by  the European integrated project AXES, the MSR/INRIA joint project and the ERC advanced grant ALLEGRO.